\documentclass[a4paper,12pt]{article}
\usepackage{geometry}
\usepackage{authblk}
\usepackage{amsmath}
\usepackage{float}
\usepackage{hyperref}
\usepackage{url}
\usepackage{graphicx}
\usepackage{xcolor}
\usepackage{pdfpages}
\usepackage{capt-of}  
\usepackage{adjustbox} 
\usepackage{afterpage} 
\usepackage{lipsum}
\title{\textsc{Exploring Next Token Prediction in Theory of Mind (ToM) Tasks: Comparative Experiments with GPT-2 and LLaMA-2 AI Models}}
\author[1]{Pavan Yadav}
\author[1]{Krishna Shinde}
\author[1]{Nikhil Khandalkar} 
\author[1]{\\Lokesh B. Ramegowda}
\author[2]{Rajarshi Das}
\affil[1]{Enkefalos Technologies} 
\affil[2]{MQube Cognition} 
\affil[1]{\texttt{\{pavan.yadav, 
krishna.shinde, nikhil.khandalkar,  lokeshbr\}@enkefalos.com}}
\affil[2]{\texttt{rajarshi.das@mqube.ai}}
\begin{document}
\date{}
\maketitle
\begin{abstract} 
\noindent Language models have significantly advanced in their ability to generate coherent text and predict subsequent tokens based on given prompts. This study systematically compares the next-token prediction performance of two widely recognized models: OpenAI’s GPT-2 and Meta’s Llama-2-7b-chat-hf in Theory of Mind (ToM) tasks. To rigorously assess their capabilities, we constructed a diverse dataset using 10 short stories sourced from the Explore ToM Dataset  \cite{ExploreToMdataset}. We enhanced these stories by programmatically inserting additional sentences (referred to as infills) using GPT-4, creating multiple variations to introduce varying levels of contextual complexity. This approach allows us to examine how increasing context influences model performance. We evaluate model behavior under different temperature settings (0.01, 0.5, 1.0, and 2.0) and test their ability to predict the next token across three distinct reasoning levels. Zero-order reasoning involves state tracking, which may probe either the current state (ground truth) or prior states  (memory). First-order reasoning refers to understanding someone’s mental state (e.g., “Does Anne know the apple is salted?”). Second-order reasoning introduces an additional level of recursion in mental state tracking (e.g., “Does Anne think that Charles knows the apple is salted?”) \cite{ExploreToMdataset}.
 Our findings reveal that increasing the number of infill sentences slightly reduces prediction accuracy, as added context introduces complexity and ambiguity. Llama-2 consistently outperforms GPT-2 in accuracy, particularly at lower temperatures, where it exhibits higher confidence in selecting the most probable next token. As question complexity increases, model responses diverge significantly. Notably, GPT-2 and  Llama-2 exhibit greater response diversity in first and second-order reasoning tasks. These insights highlight how model architecture, temperature, and context affect next-token prediction, enhancing understanding of language model capabilities and limitations.\footnote{code: \url{https://github.com/Enkefalos-Technologies/next-token-prediction}} 
\end{abstract}
 \section{Introduction}
 Language models have become an essential component of modern natural language processing (NLP), demonstrating remarkable capabilities in text generation, machine translation, and question-answering tasks. Among these, next-token prediction serves as a fundamental benchmark for evaluating model performance, as it directly influences fluency, coherence, and overall language understanding. In this study, we examine the next-token prediction capabilities of two widely recognized transformer-based models: OpenAI’s GPT-2 and Meta’s Llama-2-7b. Our goal is to analyze how these models respond under varying conditions and assess their ability to maintain accuracy across different levels of contextual complexity. We systematically evaluate the models’ performance using several experimental conditions. First, we manipulate temperature settings (0.01, 0.5, 1.0, and 2.0) to observe their impact on output randomness and confidence. Lower temperatures tend to produce more deterministic outputs, while higher temperatures introduce greater variability, potentially affecting prediction accuracy. Second, we introduce varying levels of content infill (0, 1, 4, 16, and 64 additional sentences) to assess how increasing contextual information influences token prediction. Finally, we examine the effect of question complexity by categorizing queries into zero-order, first-order, and second-order questions, each requiring different levels of reasoning and contextual dependency. 

Our findings indicate that increasing the number of infill sentences generally reduces the probability of correct token predictions, suggesting that additional content may introduce ambiguity or distract from the model’s initial response. Moreover, as question complexity increases, model responses vary significantly, highlighting differences in how each model processes and integrates contextual dependencies. Notably, both Llama-2 and GPT-2 exhibit diversity in responses, particularly in first and second-order reasoning tasks, with varying patterns in their predictions. 
Through this comparative analysis, we aim to provide insights into the strengths and limitations of these models in next-token prediction tasks. Understanding how different factors influence model behavior can guide future improvements in NLP applications, particularly in areas requiring highly accurate and context-sensitive text generation.
\newpage
\section{Related Work}
The development of language models (LMs) has progressed rapidly, enabling machines to generate coherent text, predict the next token, and perform complex reasoning tasks. A significant research focus has been on evaluating these models' ability to simulate human cognitive processes, particularly in the domain of Theory of Mind (ToM). ToM refers to the capacity to infer the mental states, beliefs, and intentions of others, a fundamental aspect of human cognition. Several recent studies have sought to assess how large language models (LLMs) approximate ToM reasoning.

Strachan et al. (2024) examined the ToM capabilities of LLMs by comparing the performance of models such as GPT-4 and LLaMA-2 against human participants on ToM tasks, including false belief attribution, indirect requests, irony detection, and faux pas recognition. Their findings suggest that while GPT-4 exhibits human-like proficiency in recognizing indirect requests, it struggles with detecting social nuances like faux pas, highlighting the complexities in evaluating ToM in LLMs~\cite{strachan2024testing}. Similarly, Das and Das (2024) introduced an iterative Theory of Mind assay to systematically evaluate AI models' reasoning abilities across multimodal inputs, providing a hierarchical assessment framework for ToM tasks~\cite{das2024iterative}.

Further research has explored next-token prediction as a key mechanism for reasoning in LLMs. Emu3, a unified multimodal transformer, demonstrated state-of-the-art performance in token-based reasoning across text, images, and video, suggesting that a single next-token prediction framework can generalize effectively to multiple modalities~\cite{emu3}. While such an approach enhances generalization, its implications for complex cognitive reasoning, such as ToM, remain underexplored.

Sclar et al. (2024) proposed an adversarial data generation approach, Explore Theory-of-Mind, to evaluate LLMs' second-order belief reasoning. Their study revealed that many LLMs struggle with hierarchical reasoning beyond direct text prediction~\cite{ExploreToMdataset}. Zhang et al. (2024) and Wang et al. (2024) further analyzed second-order belief reasoning, showing that LLMs exhibit inconsistencies in handling complex social reasoning tasks, particularly when contextual ambiguity is introduced~\cite{evaluating_theory_of_mind,to_mind_in_human_ai}. These studies underscore the need for more robust evaluation methodologies to assess LLMs' reasoning depth.

In addition to reasoning evaluation, research has focused on how LLMs process abstract representations. Hao et al. (2024) explored embedding-based reasoning, demonstrating that LLMs organize knowledge in a continuous latent space, which influences their ability to infer and generalize ToM-related tasks~\cite{training_latent_space}. Such insights provide a foundation for improving ToM modeling in AI.

Our study extends this body of work by addressing the existing gaps in ToM evaluation for LLMs. Specifically, we introduce three key modifications to assess how LLMs handle ToM reasoning under varying conditions:

\begin{itemize}
    \item \textbf{Infills:} We insert additional contextual sentences into narratives to examine whether LLMs adapt to evolving information and maintain logical consistency in their predictions. \\
    \item \textbf{Temperature Variation:} By altering the temperature parameter, we investigate how randomness in token selection impacts LLMs' ToM reasoning performance across different reasoning complexities.
    \item \textbf{Question Complexity:} We evaluate LLMs across zero-order, first-order, and second-order belief reasoning tasks to measure their ability to process increasingly complex cognitive inferences.
\end{itemize}
By integrating these factors, our study provides a more comprehensive framework for evaluating LLMs' ToM reasoning. This approach not only benchmarks current models against human cognitive capabilities but also lays the groundwork for refining AI's ability to understand and simulate social cognition. 
\section{Motivation}
As AI-driven conversational agents become increasingly integrated into real-world applications, their ability to  maintain context,  infer intent, and  reason about multiple perspectives becomes crucial. One fundamental aspect of human intelligence that supports these capabilities is  Theory of Mind (ToM)—the ability to understand and predict the thoughts, intentions, and beliefs of others.  

Our primary motivation in this study is to evaluate whether large language models (LLMs) exhibit a rudimentary form of Theory of Mind by analyzing their ability to track evolving user intentions in conversations involving multiple perspectives and contextual shifts.  

\subsection{Illustrative Example}
To highlight the importance of this issue, consider the following real-world scenario involving a ticket-booking AI assistant:
\begin{quote}
    \textbf{User:} ``I want to buy a ticket from Bengaluru to Pune."  
    
    \textbf{User:} ``Actually, my wife is considering whether we should go to Kolkata instead."  
    
    \textbf{User:} ``But I think Pune is still the best choice for me."
\end{quote}
A human assistant would naturally recognize that there are two distinct yet interrelated travel considerations—one for the user and another for the user's wife. The assistant would infer that while an alternative destination (Kolkata) was briefly considered, the user ultimately reaffirmed their preference for Pune.  

However, an AI model lacking a strong ability to track evolving intent and differentiate between conflicting perspectives might misinterpret the final decision, assuming that the user has either switched to Kolkata or remains undecided.

\subsection{Broader Implications}
This challenge extends beyond ticket booking and is particularly critical in \textbf{high-stakes domains} such as:
\begin{itemize}
    \item \textbf{Healthcare:} Misinterpretation of patient preferences could lead to incorrect treatment recommendations.
    \item \textbf{Finance:} AI-driven financial advisors must distinguish between shifting investment preferences and firm decisions.
    \item \textbf{Legal advisory:} Ambiguous client statements may result in misleading legal guidance.
\end{itemize}
If an AI system fails to \textbf{differentiate between multiple actors’ intentions} or \textbf{loses track of contextual shifts}, it risks generating \textbf{misleading or incorrect responses}, ultimately reducing its \textbf{effectiveness and trustworthiness} in real-world applications.
\subsection{Evaluating Theory of Mind via Next Token Prediction}
While LLMs do not explicitly reason like humans, their predictions are based on probability distributions over possible next words. By analyzing how these models assign probabilities to different tokens at each step, we can assess their ability to maintain context coherence, infer latent intent, and navigate conflicting perspectives. If a model consistently prioritizes contextually relevant words while predicting the next token, it suggests an implicit ability to model conversational states—a fundamental prerequisite for Theory of Mind.
\subsection{Why This Matters?}
As LLMs are increasingly deployed in customer service, insurance underwriting, climate analysis, and supply chain optimization, their ability to maintain multi-turn, multi-perspective reasoning becomes essential. This study aims to quantitatively evaluate whether state-of-the-art LLMs can infer intent in ambiguous scenarios, making them more reliable for real-world applications. By establishing a structured framework for evaluating Theory of Mind capabilities in AI, our work contributes to building the next generation of models that can reason more effectively and maintain context over extended interactions.
\section{Data Creation}
For this experiment, we constructed a dataset based on 10 short stories sourced from the ExploreToM dataset. To systematically analyze the impact of contextual complexity on next-token prediction, we generated multiple variations of each story by programmatically inserting additional sentences (referred to as infills) using GPT-4. 

\noindent Each story was modified to include five versions with varying numbers of infill sentences:
\begin{itemize}
    \item 0 infill: The original story taken from the Explore ToM dataset.
    \item 1 infill: One additional sentence inserted into the original story.
    \item 4 infills: Four additional sentences inserted.
    \item 16 infills: Sixteen additional sentences inserted.
    \item 64 infills: Sixty-four additional sentences inserted.
\end{itemize}
These infills introduce varying levels of contextual complexity, allowing us to investigate how increased information density affects the models' ability to predict the next token. Given the non-deterministic nature of large language models (LLMs), we allowed a tolerance of ±3 sentences in higher infill conditions to accommodate variations in sentence structure and coherence. \newpage
\subsection{Examples of Story Variants}
Below, we provide examples of how the original story was modified with different levels of infills:\\ \\
\textbf{0 Infill (Original Story):}
\begin{quote}
    \texttt{Kaylee entered the hotel lobby. Kaylee moved the silver letter opener to the wooden desk drawer, which is also located in the hotel lobby. While this action was happening, Liam witnessed this action in secret (and only this action). Kaylee left the hotel lobby. Liam entered the hotel lobby. Kaylee entered the hotel lobby. Liam moved the silver letter opener to the leather briefcase, which is also located in the hotel lobby.}
\end{quote}
\textbf{1 Infill (One Additional Sentence Added):} 
\begin{quote}
    \texttt{Kaylee entered the hotel lobby. Kaylee moved the silver letter opener to the wooden desk drawer, which is also located in the hotel lobby. While this action was happening, Liam witnessed this action in secret (and only this action). Kaylee left the hotel lobby. Liam entered the hotel lobby. Kaylee entered the hotel lobby. Liam moved the silver letter opener to the leather briefcase, which is also located in the hotel lobby.    \textcolor{orange}{Liam then glanced around nervously, ensuring no one else had seen him relocate the letter opener.}}
\end{quote}
\textbf{4 Infills (Four Additional Sentences Added):}
\begin{quote}
    \texttt{Kaylee entered the hotel lobby. Kaylee moved the silver letter opener to the wooden desk drawer, which is also located in the hotel lobby. While this action was happening, Liam witnessed this action in secret (and only this action). Kaylee left the hotel lobby. Liam entered the hotel lobby. Kaylee entered the hotel lobby. Liam moved the silver letter opener to the leather briefcase, which is also located in the hotel lobby.    \textcolor{orange}{Liam then glanced around nervously, ensuring no one else had seen him relocate the letter opener.
    As Liam's eyes darted across the room, he noticed a small security camera pointed directly at the briefcase.
    His heart raced as he realized that his actions might have been recorded.
    Determined to retrieve the letter opener before anyone else discovered it, Liam decided to wait in the lobby, pretending to read a newspaper while keeping a watchful eye on the briefcase.}}
\end{quote}
Similarly, 16-infill and 64-infill versions were created by adding 16 and 64 additional sentences, respectively, increasing the complexity and depth of the narrative.\newpage
\subsection{Model Evaluation Based on Question Complexity}
To systematically test model performance, we evaluated responses to three types of questions:
\begin{enumerate}
    \item Zero order questions: Basic factual retrieval questions requiring no inferential reasoning.
    \item First-order questions: Questions requiring inference based on the character’s knowledge and observed events.
    \item Second-order questions: Questions that assess the character’s belief about another character’s belief, requiring deeper reasoning.\cite{ExploreToMdataset}
\end{enumerate}
\textbf{Example Prompts:}
\begin{enumerate}
    \item \textbf{Zero-order prompt:} \texttt{"In the beginning, the silver letter opener was in the"}
    \item \textbf{First-order prompt}: \texttt{"Kaylee will look for the silver letter opener in the"} 
    \item \textbf{Second-order prompt:} \texttt{"Liam thinks that Kaylee will look for the silver letter opener in the"}
\end{enumerate}
By evaluating responses across different levels of question complexity and contextual infill, we aim to gain deeper insights into how GPT-2 and Llama-2-7b-chat-hf  handle information processing, reasoning, and next-token prediction under varying conditions.
\section{Experiment Design}
To systematically evaluate the predictive capabilities of GPT-2 and Llama-2  under different contextual complexities, we conducted next-token prediction experiments across varying levels of infill complexity. Each model was assessed under four temperature settings: 0.01 (deterministic output), 0.5 (moderate randomness), 1.0 (default setting), and 2.0 (high randomness).Each model was prompted to predict the next three tokens for each type of question, and the prompt was set as below. 
\begin{quote}
    \item \texttt{prompt = "infill type" + "zero order question"}
    \item \texttt{prompt = "Kaylee entered the hotel lobby. Kaylee moved the silver letter opener to the wooden desk drawer, which is also located in the hotel lobby. While this action was happening, Liam witnessed this action in secret (and only this action). Kaylee left the hotel lobby. Liam entered the hotel lobby. Kaylee entered the hotel lobby. Liam moved the silver letter opener to the leather briefcase, which is also located in the hotel lobby" + " " + "Liam moved the silver letter opener to the" }
\end{quote}
Across all infill levels (0, 1, 4, 16, and 64). The model's predictions were then analyzed based on probability distributions.
\subsection{Token Prediction Process}
In the token prediction sequence, the model generates a ranked list of tokens along with their corresponding probabilities. The token with the highest probability in the list is referred to as the correct token (CT), while the next four tokens are considered alternative predictions (1AP, 2AP, 3AP, 4AP) where, 1AP is the first alternative prediction, 2AP is the second alternative prediction and so on. These predictions help assess the model’s confidence and the potential alternatives considered at each step. 
\textbf{Example of Token Predictions:}
For the given prompt, we generated the next three tokens and computed their average probabilities as follows:

\begin{quote}
    \item \textbf{Step 1:}
        \begin{itemize}
            \renewcommand{\labelitemi}{}
            \item CT: `wooden', Probability: 0.8147
            \item 1AP: `hotel', Probability: 0.0681
            \item 2AP: `des', Probability: 0.0382
            \item 3AP: `front', Probability: 0.0128
            \item 4AP: `silver', Probability: 0.0085
        \end{itemize}
    \item \textbf{Step 2:}
        \begin{itemize}
            \renewcommand{\labelitemi}{}
            \item CT: `des', Probability: 0.9684
            \item 1AP: `dra', Probability: 0.0252
            \item 2AP: `hotel', Probability: 0.0045
            \item 3AP: `table', Probability: 0.0003
            \item 4AP: `desktop', Probability: 0.0002
        \end{itemize}
    \item \textbf{Step 3:}
        \begin{itemize}
            \renewcommand{\labelitemi}{}
            \item CT: `k', Probability: 1.0000
            \item 1AP: `ks', Probability: 0.0000
            \item 2AP: `in', Probability: 0.0000
            \item 3AP: `,', Probability: 0.0000
            \item 4AP: `k', Probability: 0.0000
        \end{itemize}
\end{quote}
\subsection{\textbf{Computation of Average Probabilities}} 
To quantify the model’s confidence in its predictions, we calculated the average probability for each token category (CT, 1AP, 2AP, 3AP, and 4AP) across all prediction steps. The formula for computing these averages is:
\[
\text{Average Probability} = \frac{\sum P}{N}
\]
where P represents the probability of a specific token at each step, and N is the number of steps.
Using the example predictions: 
\[
\text{Average CT Probability}: \frac{(0.8147 + 0.9684 + 1.0000)}{3} = 0.9277
\]

\[
\text{Average 1AP Probability}: \frac{(0.0681 + 0.0252 + 0.0000)}{3} = 0.0311
\]

\[
\text{Average 2AP Probability}: \frac{(0.0382 + 0.0045 + 0.0000)}{3} = 0.0142
\]

\[
\text{Average 3AP Probability}: \frac{(0.0128 + 0.0003 + 0.0000)}{3} = 0.0044
\]

\[
\text{Average 4AP Probability}: \frac{(0.0085 + 0.0002 + 0.0000)}{3} = 0.0029
\]
 \subsection{Analysis of Probability Distributions and Model Performance}
    By analyzing these probability distributions across different levels of contextual complexity and temperature settings, we aim to uncover key insights into how GPT-2 and Llama-2 process information:
    \begin{itemize}
        \item \textbf{Effect of Contextual Complexity:}
        \begin{enumerate}
            \item As the number of infills increases, does the probability of correct token prediction decrease?
            \item How does added context impact the spread of probabilities across 1AP–4AP?
        \end{enumerate}
        
        \item \textbf{Influence of Temperature Settings:}
        \begin{enumerate}
            \item Lower temperatures (0.01) should result in highly confident, deterministic predictions.
            \item Higher temperatures (2.0) introduce more randomness, leading to a flatter probability distribution across multiple tokens.
        \end{enumerate}
        
        \item \textbf{Comparative Model Performance:}
        \begin{enumerate}
            \item Does Llama-2 consistently outperform GPT-2 across different scenarios?
            \item Which model maintains higher prediction confidence when handling complex narratives?
        \end{enumerate}
    \end{itemize}
By averaging probabilities across multiple trials and contextual variations, this experiment provides a comprehensive understanding of model behavior in next-token prediction tasks. The results will help refine predictive models and optimize their contextual reasoning capabilities.

\section{Results}

The evaluation of GPT-2 and LLaMA-2 models on the given dataset revealed key differences in their performance. One notable issue was the input length constraint of the GPT-2 model, which has a maximum token limit of 1024. Due to this limitation, some questions in longer stories were skipped, whereas LLaMA-2, which supports longer contexts, was able to answer all questions without such omissions. \newline  
Additionally, as the complexity of questions increased, the average probability of correct responses slightly decreased for both models. This suggests that the models struggled more with nuanced or multi-step reasoning questions compared to simpler ones.  
Another key observation was the change in model responses for CT (Correct Tokens). The variations in GPT-2’s responses compared to LLaMA-2 highlight potential differences in how these models handle contextual understanding, reasoning depth, or specific knowledge areas.  
These findings underscore the impact of input length constraints, question complexity, and model-specific response patterns, providing insights into their strengths and limitations for next-token prediction and related tasks.

When analyzing the responses at different temperature settings, we observed that as the temperature increased, both GPT-2 and LLaMA-2 exhibited a decline in prediction probability, indicating reduced confidence in their responses. Additionally, higher temperatures led to greater response diversity, with both models generating a wider range of completions. However, this also increased the likelihood of incorrect or less coherent outputs. While GPT-2 displayed more variability in its responses across temperature changes, LLaMA-2 maintained relatively more consistent predictions, suggesting a stronger ability to retain contextual coherence despite increased randomness in token selection.

A detailed examination of zero-order, first-order, and second-order questions further highlighted the impact of increasing temperature and complexity on model performance. Zero-order questions, which required direct factual recall, were answered with the highest accuracy by both models. As the complexity of reasoning increased to first-order questions—requiring inferences based on character knowledge and observed events—the prediction probability showed a moderate decline. The most significant drop was observed in second-order questions, which required reasoning about a character’s belief regarding another character’s belief. The increased challenge of these questions, combined with higher temperatures, led to more diverse responses but also a greater probability of errors.

Similarly, when analyzing context infill, both models showed a decline in prediction probability as the number of missing tokens increased. This trend suggests that the difficulty of accurately completing a sentence or passage grows with longer gaps, making the models less certain about their predictions. At the same time, both models demonstrated greater response diversity as infill length increased, producing multiple plausible completions rather than converging on a single high-confidence answer.

Overall, our findings indicate that while increasing temperature and infill length enhance the diversity of model outputs, they also reduce prediction confidence, leading to a higher probability of incorrect or incoherent responses. These trade-offs highlight the challenges of balancing diversity and accuracy in next-token prediction tasks.

\subsection{GPT-2 Model, Increasing Temperature, and Increasing Infills for Second Order Prompt}

The following plots illustrate the results for the second-order question from the first story in the ExploreToM dataset, based on the prompt:  
\texttt{"Liam thinks that Kaylee will look for the silver letter opener in the"}  

These results were generated using the GPT-2 model with a increasing temperature and increasing infills.
The rest of the results are presented in Appendix~\ref{appendix:results}
\begin{figure}[H] 
    \centering
    \includegraphics[width=\textwidth]{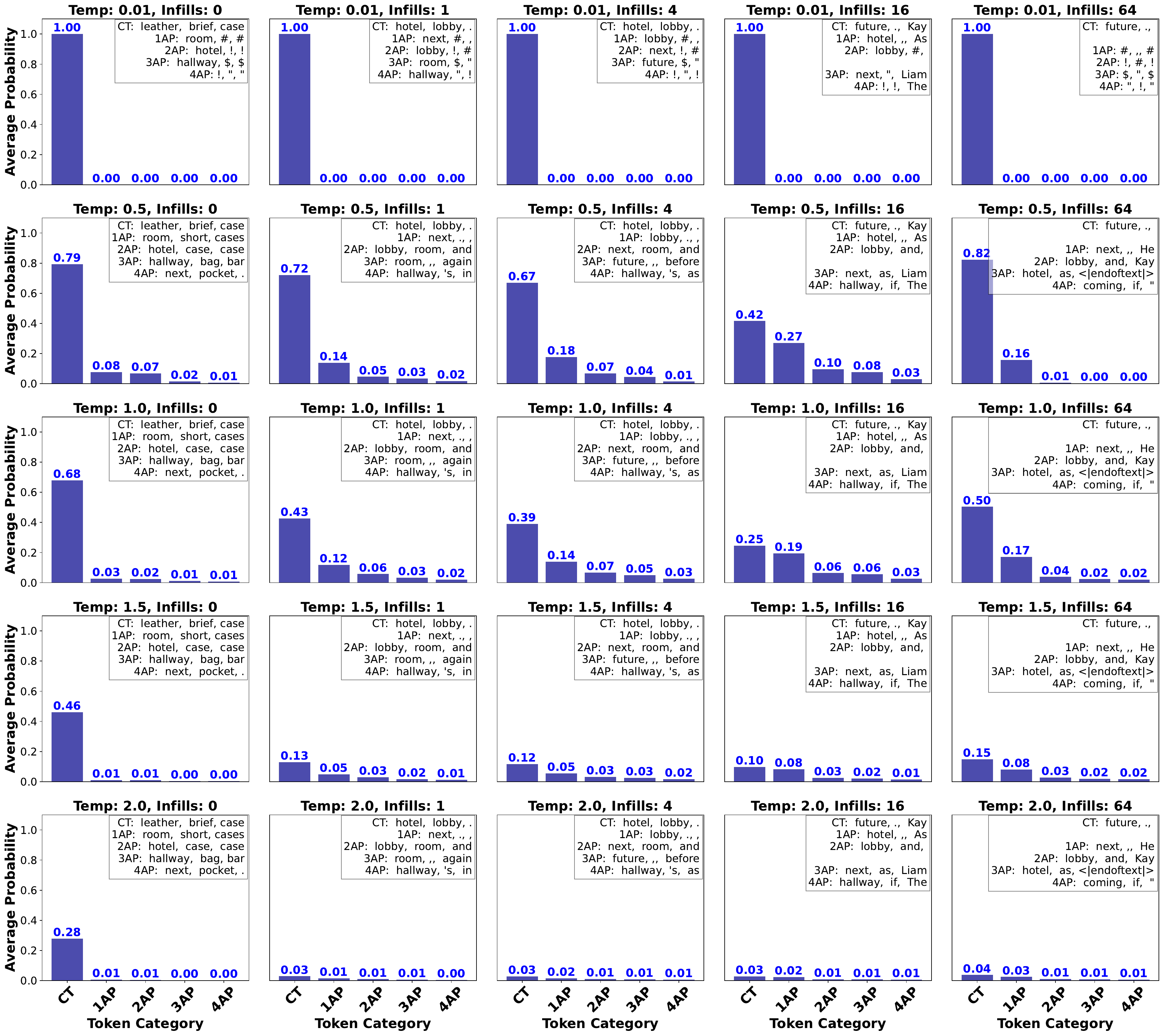}
    \caption{Prediction results of the GPT-2 model for different temperature settings and infill variations.}
    \label{fig:gpt2_results}
\end{figure}\newpage
\subsection{Llama-2 Model, Increasing Temperature, and Increasing Infills for Second Order Prompts}

The following plots illustrate the results for the second-order question from the first story in the ExploreToM dataset. The prompt used was:  
\texttt{"Liam thinks that Kaylee will look for the silver letter opener in the"}  

These results were generated using the Llama-2 model with a temperature setting of 0.01, 0.5, 1.0, 1.5, and 2.0 while progressively increasing the number of infills. The experiment aims to assess how varying temperature and added contextual information affect the model’s predictive accuracy. The rest of the results are presented in Appendix~\ref{appendix:results}  

    \begin{figure}[H] 
    \centering
    \includegraphics[width=\textwidth]{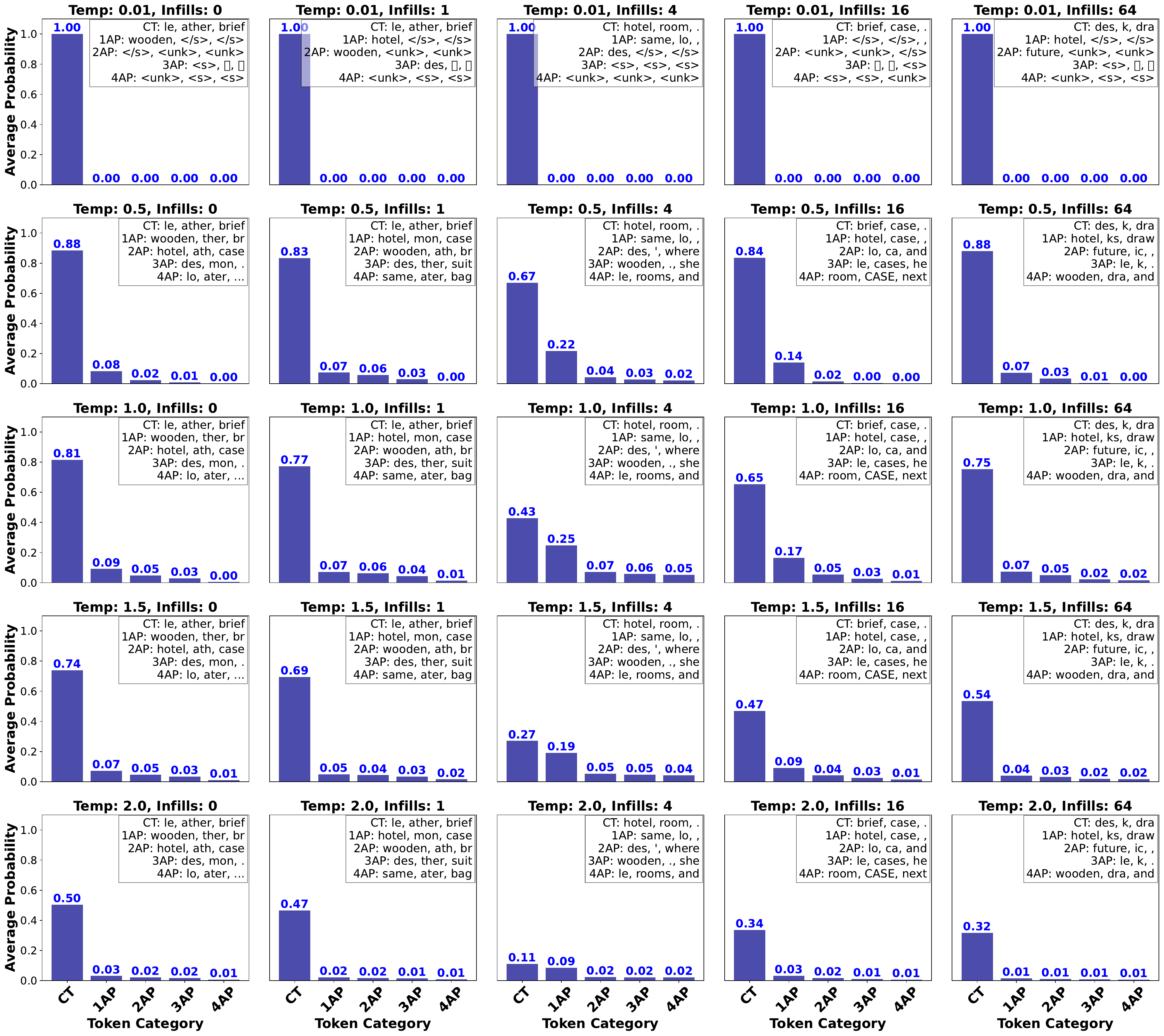}
    \caption{Prediction results of the Llama-2 model for different temperature settings and infill variations.}
    \label{fig:llama2_results}
    \end{figure}
\newpage
\section{Conclusion}
This study provides a comparative analysis of next-token prediction performance between \texttt{GPT-2}  and \texttt{Llama-2-7b-chat-hf} under varying contextual complexity, temperature settings, and question types. The results indicate that increasing infill density reduces prediction accuracy for both models, though \texttt{Llama-2} demonstrates greater resilience compared to \texttt{GPT-2}. Temperature settings also play a significant role, with lower temperatures leading to more deterministic and accurate predictions, whereas higher temperatures introduce variability at the expense of accuracy.

In terms of question complexity, \texttt{Llama-2} outperforms \texttt{GPT-2} in higher-order reasoning tasks, particularly in second-order inference questions. Additionally, while \texttt{Llama-2} achieves superior accuracy, \texttt{GPT-2} remains a faster alternative, making it more suitable for real-time applications where computational efficiency is a priority.

These findings contribute to a broader understanding of the strengths and limitations of different transformer architectures in next-token prediction. Future research may focus on fine-tuning strategies to enhance model robustness under complex contextual conditions and further evaluate trade-offs in real-world NLP applications.
\section{Future Work}

Our future research aims to understand the internal mechanisms of transformer-based language models by exploring both interpretability techniques and computational tools. Specifically, we seek to analyze how embeddings and hidden states evolve during prediction to gain deeper insights into how contextual information is processed at different layers. To achieve this, we are developing interpretability techniques that enhance transparency in the decision-making process of large language models. Additionally, studies on world models in LLMs suggest that these models construct internal representations of their environment, providing valuable insights into how token prediction mechanisms operate in complex linguistic settings~\cite{aiguide_part1, aiguide_part2}. To support this research, we are integrating computational tools such as Sparse Autoencoder (SAE) visualization and TransformerLens, which have shown promise in uncovering interpretable features by identifying sparse activations correlated with specific linguistic structures~\cite{transformerlens}. These tools will enable fine-grained analyses of token representations, bridging the gap between raw model activations and human-understandable reasoning patterns.  

Furthermore, recent work on mechanistic interpretability suggests that attribution graphs and circuit-level analyses within transformer models can reveal meaningful substructures responsible for specific computations~\cite{arxiv_2410_21272,transformer_circuits}. Leveraging these insights will allow us to better understand how different model components contribute to prediction tasks.  

Overall, this research will advance mechanistic interpretability in language models, laying the groundwork for refining, interpreting, and improving next-token prediction models.  


\appendix
\section{Appendix}
\label{appendix:results} 
In this appendix, we present a series of supplementary materials related to the experiments and results discussed in the main body of the document. Specifically, we include the full set of PDFs that showcase the outputs of various models under different conditions. These outputs provide insights into the performance of models across different question types and their behavior.

Each figure in the following pages corresponds to a specific model (e.g., GPT-2 or LLaMA-2) and demonstrates its performance for different orders, such as Zero Order, First Order, and Second Order. These orders represent varying levels of complexity or input types, allowing for a deeper analysis of how the models respond to different stimuli.
The following sections contain the PDF outputs for each model:

\begin{itemize}
    \item \textbf{GPT-2 Zero Order}: Results demonstrating the GPT-2 model's performance on Zero Order tasks with increasing infills and increasing temperature.
    \item \textbf{GPT-2 First Order}: Outputs showing the GPT-2 model's behavior with First Order tasks with increasing infills and increasing temperature.
    \item \textbf{GPT-2 Second Order}: Performance analysis of the GPT-2 model under Second Order questions with increasing infills and increasing temperature.
    \item \textbf{LLaMA-2 Zero Order}: LLaMA-2 model results for Zero Order tasks with increasing infills and increasing temperature.
    \item \textbf{LLaMA-2 First Order}: Performance of LLaMA-2 under First Order questions with increasing infills and increasing temperature.
    \item \textbf{LLaMA-2 Second Order}: Evaluation of LLaMA-2 on Second Order question with increasing infills and increasing temperature.
\end{itemize}

These results contribute to the broader analysis of model capabilities and provide a detailed comparison across different architectures and task complexities. 

\newcounter{storyCounter}

\newcommand{\includePDFWithStory}[3]{%
    \setcounter{storyCounter}{1} 
    \includepdf[pages=1-,scale=0.85,offset=0mm 35mm,pagecommand={%
        \centering
        \vspace{-5mm} 
        \textbf{\small #1 Story: \thestoryCounter, Prompt Type: #2}
        \vspace{-10mm} 
        \stepcounter{storyCounter}
    }]{#3}
}
\includePDFWithStory{Model: GPT-2,}{Zero Order}{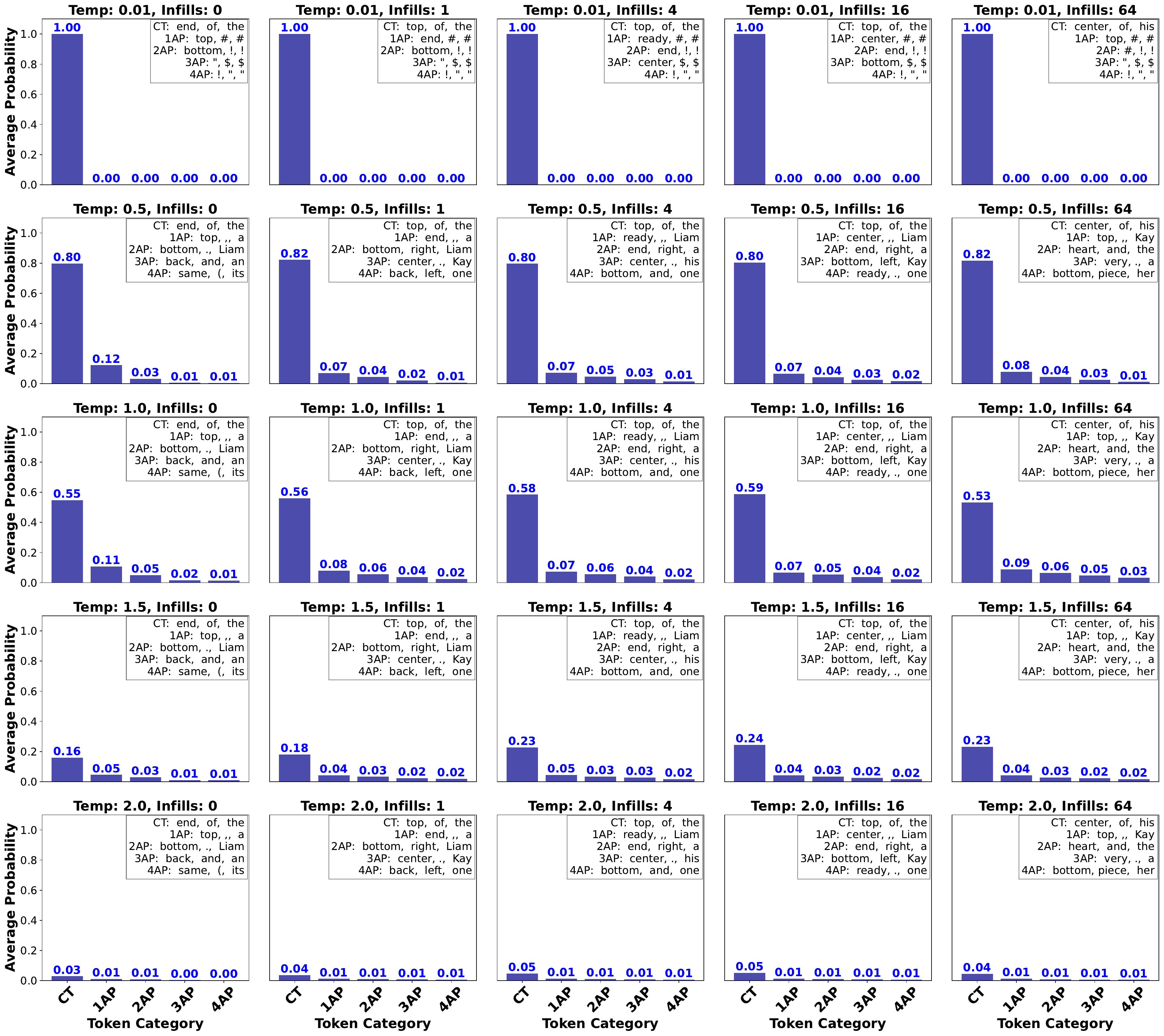}
\includePDFWithStory{Model: GPT-2,}{First Order}{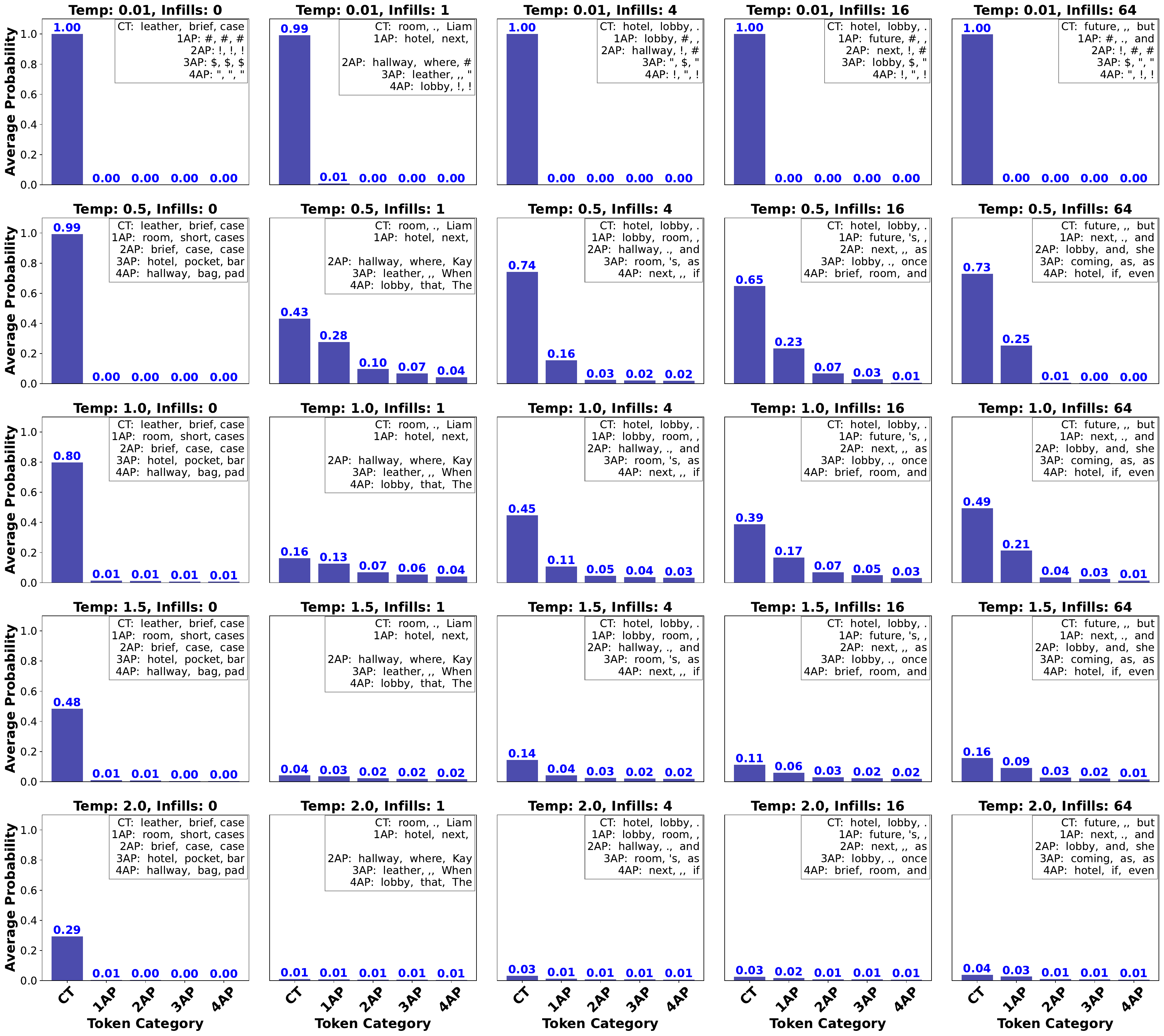}
\includePDFWithStory{Model: GPT-2,}{Second Order}{gpt2_second_order_result.pdf}
\includePDFWithStory{Model: LLAMA-2,}{Zero Order}{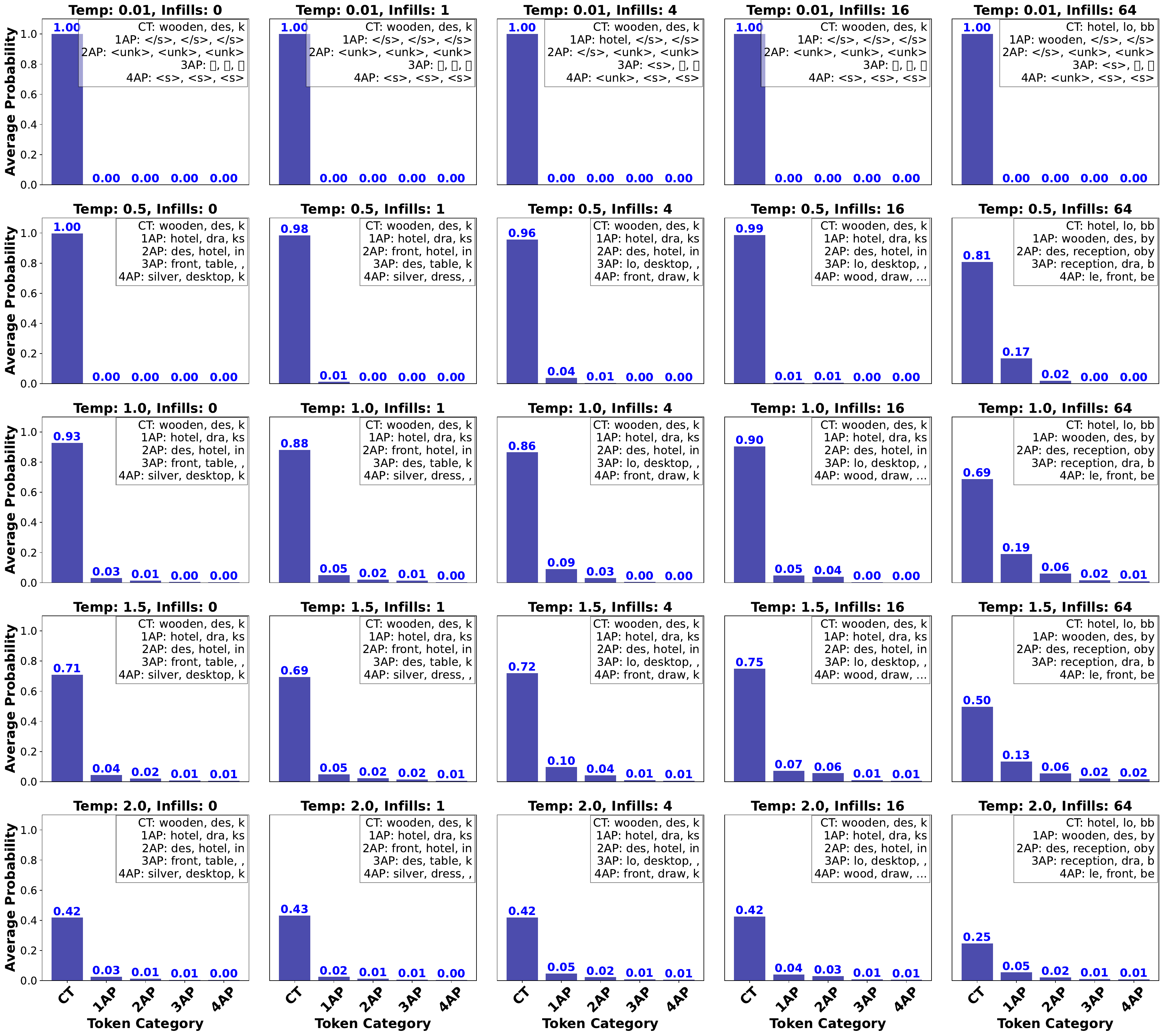}
\includePDFWithStory{Model: LLAMA-2,}{First Order}{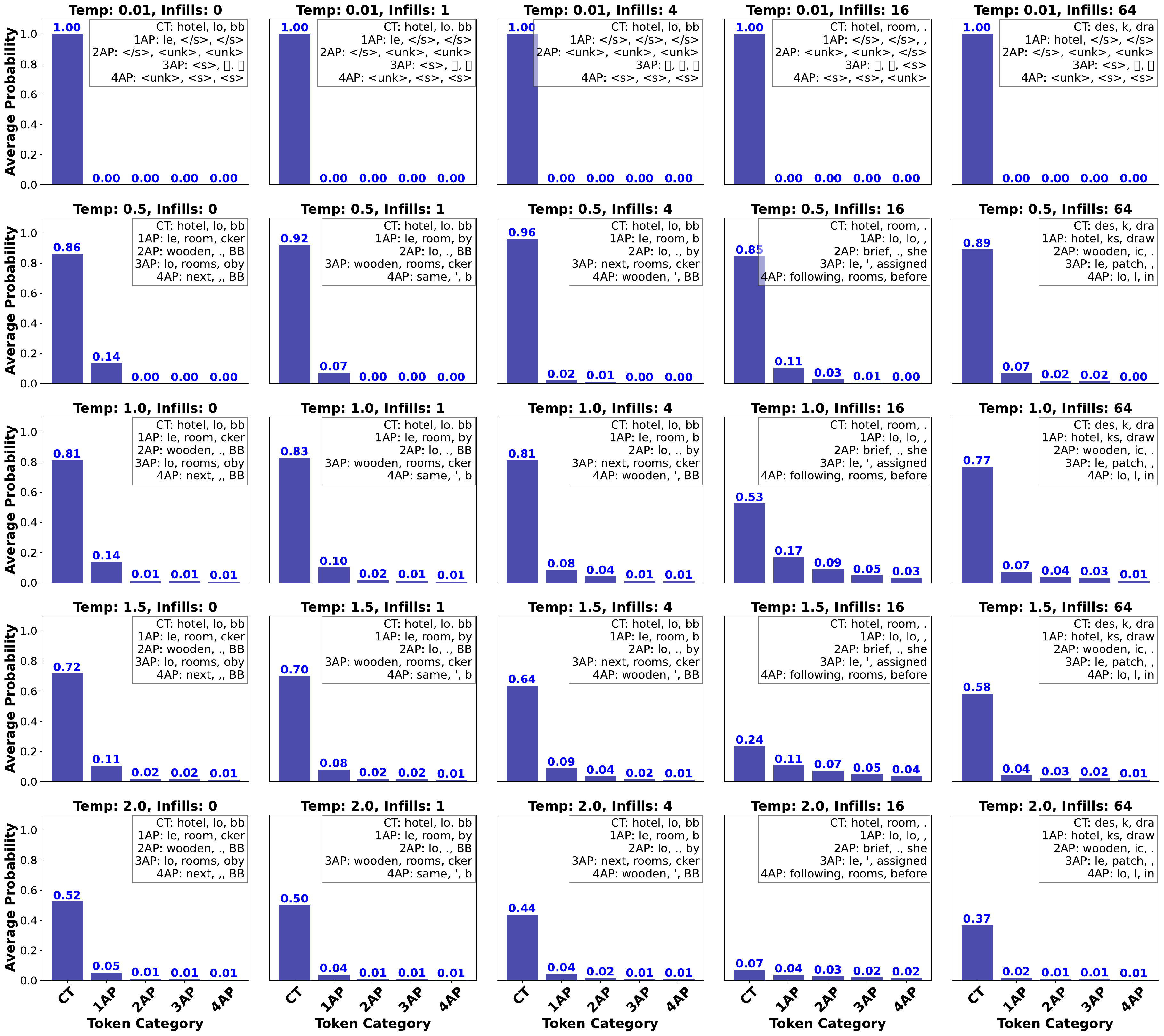}
\includePDFWithStory{Model: LLAMA-2,}{Second Order}{llama2_so_result.pdf}

\begin{thebibliography}{99}
\bibitem{ExploreToMdataset}
Sclar, M., Yu, J., Fazel-Zarandi, M., Tsvetkov, Y., Bisk, Y., Choi, Y., \& Celikyilmaz, A. (2024). \textbf{Explore Theory-of-Mind: Program-Guided Adversarial Data Generation for Theory of Mind Reasoning}. arXiv preprint. \url{https://doi.org/10.48550/arXiv.2412.12175}
\bibitem{strachan2024testing}
Strachan, M., et al. (2024). \textbf{Testing the Theory of Mind Capabilities of Large Language Models}. Nature, 627, 2024. \url{https://www.nature.com/articles/s41562-024-01882-z}

\bibitem{emu3}
Emu3 Team. (2024). \textbf{Emu3: Next-Token Prediction is All You Need}. arXiv preprint. \url{https://arxiv.org/pdf/2409.18869}
\bibitem{evaluating_theory_of_mind}
Zhang, W., Qian, Y., Cao, X., \& Shi, S. (2024). \textbf{Evaluating Theory-of-Mind in Large Language Models: A Case Study on Second-Order Belief Reasoning}. arXiv preprint. \url{https://doi.org/10.48550/arXiv.2412.06769}


\bibitem{llama2}
Meta. \textbf{Meta’s Llama-2 Model: meta-llama/Llama-2-7b-chat-hf}. Hugging Face. \url{https://huggingface.co/meta-llama/Llama-2-7b-chat-hf}

\bibitem{gpt2}
OpenAI. \textbf{OpenAI’s GPT-2 Model: GPT-2 on Hugging Face}. Hugging Face. \url{https://huggingface.co/gpt-2}

\bibitem{plots}
\textit{Saved Plots and Supplementary Material}, \url{https://github.com/Enkefalos-Technologies/next-token-prediction}


\bibitem{das2024iterative}  
Das, R. E., \& Das, R. (2024). \textit{Iterative Theory of Mind Assay of Multimodal AI Models}.  
In \textit{ICML 2024 Workshop on LLMs and Cognition}.  
\url{https://openreview.net/forum?id=PsGVVQJZGk}

\bibitem{training_latent_space}
Hao, S., Sukhbaatar, S., Su, D., Li, X., Hu, Z., Weston, J., \& Tian, Y. (2024). \textbf{Training Large Language Models to Reason in a Continuous Latent Space}. arXiv preprint. \url{https://doi.org/10.48550/arXiv.2412.06769}

\bibitem{llm_abstraction_reasoning}
Mitchell, M., \& Krakauer, D. C. (2023). \textbf{The debate over understanding in AI’s large language models}. Proceedings of the National Academy of Sciences, 120(13). \url{https://doi.org/10.1073/pnas.2215907120}

\bibitem{to_mind_in_human_ai}
Wang, Q., Walsh, S., Si, M., Kephart, J., Weisz, J., \& Noel, A. (2024). \textbf{Theory of mind in human–AI interaction}. In Extended Abstracts of the CHI Conference on Human Factors in Computing Systems. \url{https://doi.org/10.1145/3613905.3636308}


\bibitem{aiguide_part1}
Melanie Mitchell, AI Guide. (2024). \textbf{LLMs and World Models - Part 1}. Substack. \url{https://aiguide.substack.com/p/llms-and-world-models-part-1}

\bibitem{aiguide_part2}
Melanie Mitchell, AI Guide. (2024). \textbf{LLMs and World Models - Part 2}. Substack. \url{https://aiguide.substack.com/p/llms-and-world-models-part-2}

\bibitem{transformerlens}
Nanda, N., et al. (2024). \textbf{TransformerLens: Interpretability for Transformer Models}. GitHub. \url{https://transformerlensorg.github.io/TransformerLens/}
\bibitem{arxiv_2410_21272}
Yaniv Nikankin. (2024). \textbf{Arithmetic Without Algorithms: Language Models Solve Math with a Bag of Heuristics
}. arXiv preprint. \url{https://arxiv.org/pdf/2410.21272}

\bibitem{transformer_circuits}
Emmanuel A., et al. (2025). \textbf{Circuit Tracing: Revealing Computational Graphs in Language Models}. \url{https://transformer-circuits.pub/2025/attribution-graphs/methods.html}
\end{thebibliography}
 \end{document}